%% file: sample.tex
\documentclass[pmlr]{jmlr}

\makeatletter
\def\set@curr@file#1{\def\@curr@file{#1}} 
\makeatother
\usepackage[load-configurations=version-1]{siunitx} 

\usepackage{lipsum}
\usepackage{mwe}

\usepackage{tikz}

\makeatletter
\newcommand*{\centernot}{%
  \mathpalette\@centernot
}
\def\@centernot#1#2{%
  \mathrel{%
    \rlap{%
      \settowidth\dimen@{$\m@th#1{#2}$}%
      \kern.5\dimen@
      \settowidth\dimen@{$\m@th#1=$}%
      \kern-.5\dimen@
      $\m@th#1\not$%
    }%
    {#2}%
  }%
}
\makeatother

\usepackage{mathtools}

\DeclarePairedDelimiterX\Basics[1](){ #1}

\newtheorem{assumption}{Assumption}

\usepackage{algorithm}
\usepackage{algorithmic}
\usepackage{enumitem}
\usepackage{multirow}

\usepackage{longtable}

\usepackage{booktabs}
\usepackage[load-configurations=version-1]{siunitx} 

\usepackage{lscape}
\usepackage{multicol}
\usepackage{hyperref}

\theorembodyfont{\upshape}
\theoremheaderfont{\scshape}
\theorempostheader{:}
\theoremsep{\newline}

\jmlrvolume{XXX}
\jmlryear{2022}
\jmlrworkshop{XXXX}

\title[Causal Knowledge Hierarchy for Structural Causal Models]{CKH: Causal Knowledge Hierarchy for Estimating Structural Causal Models from Data and Priors}

\author{%
\Name{Riddhiman Adib} \Email{adib@ohsu.edu}\\
\addr Oregon Clinical and Translational Research Institute, Oregon Health \& Science University
\AND
\Name{Md Mobasshir Arshed Naved} \Email{naved@purdue.edu}\\
\addr Department of Computer Science, Purdue University
\AND
\Name{Chih-Hao Fang} \Email{fang150@purdue.edu}\\
\addr Department of Computer Science, Purdue University
\AND
\Name{Md Osman Gani} \Email{mogani@umbc.edu}\\
\addr Department of Information Systems, University of Maryland, Baltimore County
\AND
\Name{Ananth Grama} \Email{ayg@cs.purdue.edu}\\
\addr Department of Computer Science, Marquette University
\AND
\Name{Paul Griffin} \Email{pmg14@psu.edu}\\
\addr Department of Industrial Engineering, Pennsylvania State University
\AND
\Name{Sheikh Iqbal Ahamed} \Email{sheikh.ahamed@marquette.edu}\\
\addr Department of Computer Science, Marquette University
\AND
\Name{Mohammad Adibuzzaman} \Email{adibuzza@ohsu.edu}\\
\addr Oregon Clinical and Translational Research Institute, Oregon Health \& Science University
}

\begin{document}

\maketitle


\input{sections/0_abstract}
\input{sections/1_introduction}
\input{sections/2_background} 
\input{sections/3_method} 
\input{sections/algorithm}
\input{sections/4_simulation} 
\input{sections/5_discussion} 
\input{sections/6_acknowledgments}


\bibliography{sample}

\input{sections/8_appendix}


\end{document}

%% file: sections/0_abstract.tex
\begin{abstract}
    Structural causal models (SCMs) provide a principled approach to identifying causation from observational and experimental data in disciplines ranging from economics to medicine. However, SCMs, which is typically represented as graphical models, cannot rely only on data, rather require support of domain knowledge. A key challenge in this context is the absence of a methodological framework for encoding priors (background knowledge) into causal models in a systematic manner. We propose an abstraction called causal knowledge hierarchy (CKH) for encoding priors into causal models. Our approach is based on the foundation of ``levels of evidence" in medicine, with a focus on confidence in causal information. Using CKH, we present a methodological framework for encoding causal priors from various information sources and combining them to derive an SCM. We evaluate our approach on a simulated dataset and demonstrate overall performance compared to the ground truth causal model with sensitivity analysis.
\end{abstract}

%% file: sections/1_introduction.tex
\section{Introduction}
Causal knowledge discovery or identifying cause-and-effect relationships between variables is a fundamental objective in various domains such as epidemiology and medicine \citep{rothman2005causation}, sociology \citep{gangl2010causal}, and economics \citep{gow2016causal}. Without understanding causal relationships, scientists rely on correlations, which do not allow for estimation of intervention effect (i.e., doing) of a variable on a model outcome. While randomized controlled trials remain the gold standard for exploring causation \citep{kovesdy2012observational}, they are often infeasible because of cost, time, and/or ethical reasons \citep{glass2013causal}. Thus, causal discovery from observational data that is complementary to experimental studies is of significant interest \citep{hernan2018c, listl2016causal, nichols2007causal}. 

Recent developments in the theory of causal inference under the Pearl causal hierarchy (PCH) \citep{bareinboim2020pearl,pearl2016causal,pearl2009causal,bareinboim2016causal}, also known as structural theory of causation (within the potential outcome framework) provides the methodologies to estimate causal effects from observational data. Within this, a causal model is expressed through structural causal models (SCMs) \citep{pearl2010causal}. SCMs represent variables of interest (exogenous and endogenous), causal relationships between the variables, and underlying probability distributions. SCMs use a graphical representation of the causal model, formalize the identification of causal effects from observational and experimental data, estimate the interventional distribution $(P(y|do(x)))$ through do-calculus \citep{pearl2019seven}, and assess hypothetical scenarios from available data and model with necessary assumptions explicitly encoded into the model.

PCH is grounded in the three layers of causation -- ($L_1$) seeing, ($L_2$) doing, and ($L_3$) imagining. Recent work on the Pearl causal hierarchy (PCH) proved that discovery in a higher layer of causation using only information from a lower layer is not feasible \citep{bareinboim2020pearl}. In other words, estimating the effect of experimentation (i.e., ``doing ($L_2$)") is not feasible based only on observational data (i.e., ``seeing ($L_1$)" ) \citep{bareinboim2020pearl}. Hence it is critical to augment observational data ($L_1$) with other sources of information (e.g., expert knowledge) to derive the effect of intervention. Expert knowledge can come in different forms, such as, expert opinions, established causal relationships and, peer-reviewed literature \citep{druzdzel2003combining}. Within each of these forms, confidence in knowledge can vary based on the knowledge source and acceptability by scientific community. However, no methodological framework exists for incorporating domain expertise with data-driven causal discovery from observational data in a systematic way
\citep{spirtes2016causal}.

In this work, we develop a methodological framework to augment data-driven causal discovery tools with human in the loop (HTL) models. The additional causal knowledge sources include different tiers of knowledge, but is not limited to: background knowledge, expert opinion, and literature. For this purpose, we have broadly categorized possible causal knowledge sources into three tiers and proposed a causal knowledge hierarchy (CKH) between the tiers. Using this causal knowledge hierarchy, we develop an associated standardized methodology to curate necessary causal information and merge them to derive the structural causal model (SCM). We also provide both theoretical and simulated results of the framework along with algorithmic pseudocode detailing the implementation.

We make the following specific contributions in this work:
\begin{enumerate}
    \item We propose a causal knowledge hierarchy (CKH) on the foundation of levels of evidence in medicine based on the confidence in the causal information.
    \item We present a standard methodological pipeline, based on CKH, to capture causal knowledge from different sources and combine them to derive the SCM.
    \item We show the effectiveness of our proposed method in a simulated experiment, detailing the implementation, along with evaluation. 
\end{enumerate}

%% file: sections/2_background.tex
\section{Related Work}

The need for causal inference from sources other than data has been explored in the literature \citep{spirtes2016causal}. From a theoretical perspective, research has been conducted on integrating causal information from varying sources. \citet{lee2020causal} proposed GID-PO that identifies the causal effect from partially-observed distributions. In another result \citep{lee2020general}, an algorithmic approach that combines data collected under multiple, disparate regimes (observational and interventional) to identify specific causal effects was presented. However, the focus of these experiments was not on prior knowledge or varying knowledge sources. \citet{borboudakis2012incorporating} used path-constraints to incorporate prior causal knowledge, without explicitly discussing the impact of causal knowledge sources. For tiered knowledge, \citet{andrews2020completeness} proposed tiered background knowledge where each tier consists of sets of variables with causal relationships, preceding another set of variables (aka, tier), and demonstrated that FCI (Fast Causal Inference) is a sound and complete causal structure elarning algorithm with and incorporation of this knowledge. From an empirical viewpoint, \citet{nordon2019building} developed a causal model from medical literature and electronic medical record (EMR) data, by generating two independent graphs, one based on the literature and one from the EMR data, and merged them. The method did not consider other sources of knowledge and did not compare knowledge sources as well as their confidence of information. In a related result, a prior-knowledge-based causal discovery algorithm \citep{wang2020prior} has been proposed to discover the underlying causal mechanism between bone mineral density and its factors from the clinical data, where prior knowledge was handpicked manually and added to the algorithm as a whitelist between edges. Finally, from a software application perspective, a graphically similar software application to help researchers navigate published findings has been proposed by the software ``ResearchMaps" \citep{landreth2013need}. However, this is primarily a visualization tool to illustrate interconnected features with a graph. In summary, although previous research has attempted to resolve causal information from varying sources, a unified and principled approach to build a generic SCM is still needed.

%% file: sections/3_method.tex
\section{Expert Augmented Causal Model with Knowledge Hierarchy}
\begin{figure*}[ht]
    \centering
    \includegraphics[width=\textwidth]{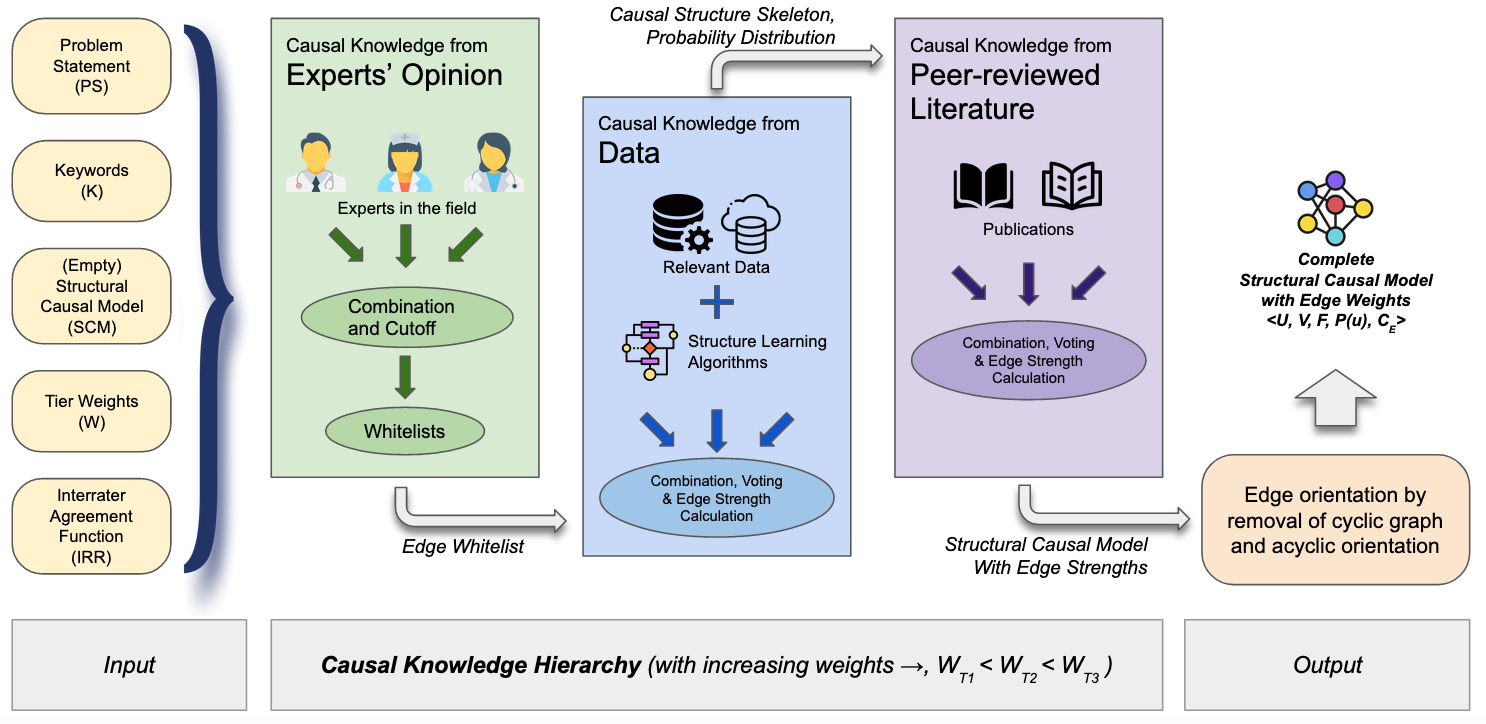}
    \caption{Our approach starts with the inputs: problem statement (PS), defined keywords (K), (empty) structural causal model (SCM), pre-defined tier weights (W), and Inter-rater agreement function (IRR) (\textit{discussed in supplementary document}). For each tier, a general series of steps is described. Finally, the SCM goes through an edge orientation phase to produce the fully specified SCM with individual edge weights ($\langle U,V,F,P(u),C_E \rangle$)}
    \label{fig:process_summary}
\end{figure*}

We propose the concept of tiers of causal knowledge and the generalized algorithm for causal model learning through knowledge hierarchy. Specifically, we propose \textit{Causal Knowledge Hierarchy (CKH)} that uses three tiers of knowledge analogous to the ``hierarchy of evidence" \citep{ackley2008evidence} and the associated weight for each tier. We discuss the assumptions within the CKH required for our proposed methodology. Finally, we establish step-by-step actions for the method. An overview of our method is presented in \autoref{fig:process_summary}. 

\subsection{Causal Knowledge Hierarchy}
Levels of evidence \citep{ackley2008evidence}, is a well established knowledge hierarchy based on the study design, data collection, and sample size. Based on this concept we propose a ``Causal Knowledge Hierarchy (CKH)" to incorporate causal information from different sources. 

CKH is a multi-level descriptor between types of knowledge and their contribution to the overall causal structure in a problem domain. We initially define three common sources of causal information and propose a hierarchy between them. We define necessary assumptions to make our proposed framework effective and generalizable. We categorize sources of causal knowledge for scientific studies into three distinct classes and define a hierarchy (\textbf{CKH}) based on the statistical confidence in the causal information they hold. 

\begin{definition}
  The three tiers of the CKH are: (1) \textbf{Tier 1: Causal knowledge from expert opinion / expertise ($\mathbf{CK_E}$)}, (2) \textbf{Tier 2: Causal knowledge from data ($\mathbf{CK_D}$)}, and (3) \textbf{Tier 3: Causal knowledge from peer-reviewed literature ($\mathbf{CK_L}$)}. The target structural causal model is a function of convex combination of causal knowledges from the three tiers of sources, $$SCM \gets f(CK_E, CK_D, CK_L)$$
\end{definition}

\begin{figure}[ht]
    \centering
    \includegraphics[width=0.3\textwidth]{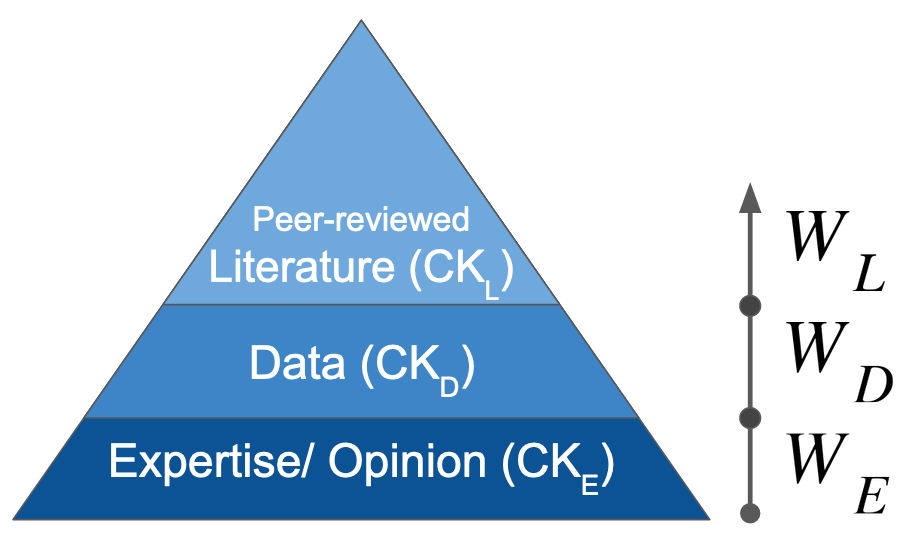}
    \caption{Tiers of Causal Knowledge Source}
    \label{fig:tiers}
\end{figure}



\subsubsection{Tier 1: Causal Knowledge from Expert Opinion}
Tier 1 of CKH incorporates causal knowledge based on the expertise and opinions ($\mathbf{CK_E}$) from researchers, scientists, and, subject matter experts (SME). This includes, but is not limited to, inputs from physicians, discussion with application users and intervention participants and, by researchers working in a specific problem domain; and excludes any knowledge directly from peer-reviewed literature. Causal knowledge is generally captured through surveys or structured communications' methods-driven group discussions (e.g., Delphi method \citep{linstone1975delphi}). This collaborative knowledge requires further validitation through scientific studies, and is prone to high levels of bias due to variation in the expert's training and experience. We classify this causal knowledge as Tier 1 ($\mathbf{CK_E}$), and assign a lower weight ($\mathbf{W_E}$) since it contributes diverse information with lesser confidence.

\subsubsection{Tier 2: Causal Knowledge from Data}

Tier 2 encodes causal knowledge generated from data sources ($\mathbf{CK_D}$). Data can be from various study designs such as an experimental study (e.g., data from randomized controlled trials), an observational study (e.g., text mining data from social media), or in between (e.g., data from pragmatic clinical trial). Depending on the data generation mechanism, different structural causal models are used to explain the causal relationships between the variables. However, there may be bias from selection, confounding, or other experimental design features. We associate the causal knowledge gathered from data ($\mathbf{CK_D}$) at Tier 2, with a relatively higher weight ($\mathbf{W_D}$) than that of Tier 1 ($\mathbf{CK_E}$). The rationale for this is: (a) data can be collected from different study designs, locations, and corroborated over time, (b) data can be analyzed further with newer methodologies and models, and (c) data can convey the effect of causal relationships between covariates for scientific studies. Because of the higher weights, it contributes more to that conjoined causal model, and can even alter directions of certain causal relationships defined from Tier 1.

\subsubsection{Tier 3: Causal Knowledge from Literature}

Tier 3 is causal information from peer-reviewed literature ($CK_L$) and has the highest weight in the CKH. It excludes any knowledge from opinions of experts, without references. Examples of Tier 3 include causal knowledge from peer-reviewed and published literature, systematic reviews, meta-analyses, evidence syntheses, article synopses, and causal effect of interventions published as studies. Within Tier 3, there may be different levels of evidence. The data extraction process additionally falls under the domain of text-mining and natural language processing (NLP). Causal knowledge from literature ($\mathbf{CK_L}$) may have its own biases such as selection bias for inclusion exclusion criteria or transportability bias for differences in population.





\subsubsection{Tier Weights for Causal Knowledge Hierarchy}

\begin{axiom}
    Each tier of causal knowledge hierarchy (CKH) has individual weights ($W_E, W_D, W_L$), signifying the confidence of the causal information. A higher tier holds a higher weight and provides more robust causal information compared to that from lower tiers. By definition, $$CK_E \propto W_E, \; CK_D \propto W_D, \; CK_L \propto W_L$$
\end{axiom}
\vspace{-2mm}
Based on the causal information hierarchy proposed, we define three weights $W = \{W_E, W_D, W_L\}$ for each tier of the causal knowledge hierarchy (\textit{refer to Axiom 1}). The weights are defined such that: 
\begin{enumerate}[nolistsep]
    \item $\mathbf{\sum_{i} W_{i} = 1}$: By definition of convex combination, the sum of all three tier weights ($\sum W = W_E + W_D + W_L$) is $1.0$. A full agreement for a specific edge connection and direction from all three tiers of the CKH results in maximum edge confidence of $1.0$.
    \item $\mathbf{W_E < W_D < W_L}$: The weights are defined in an increasing order. At any time, causal information from one tier can only contribute a maximum of their tier weight. Thus, this increasing score ensures a hierarchy between each tier. 
\end{enumerate}


\subsubsection{Assumptions}

For our proposed methodological framework, there are two associated assumptions. 

\begin{assumption}
    Knowledge within the same tier of CKH does not override one another.
\end{assumption}

For conflict resolutions with contradictory causal information within the same tier (such as, $A \rightarrow B$ from dataset 1 and $A \leftarrow B$ from dataset 2, both from $CK_D$),we find the strength of the causal relationship based on all the information within the same tier. Selection of knowledge sources within a tier is subjective and depends on the experimenter. Consequently, we do not propose any hierarchy within a tier, rather, we compute the conjoined strength of the causal connections. A similar direction in causal connections and edges increase the confidence, whereas contradictory causal connections and edges reduce the confidence of the edge. 

\begin{assumption}
    Within CKH, knowledge from upper tier (or, in special case, tier with more weight) can reverse/ override knowledge from lower tier.
\end{assumption}

Unlike the earlier assumption, when we have contradictory causal information from different tiers (such as, $A \rightarrow B$ from Tier 1 ($CK_E$) and $A \leftarrow B$ from Tier 2 ($CK_D$)), the direction of causal relationship from an upper tier can override that from a lower tier. 

Our assumption on brief classification of CKH is presented in \autoref{tab:ckh_basic}, based on how many individual has worked for the knowledge curation and validation. The higher the number of people worked on establishing the knowledge, the more weight it holds in CKH; simply based on the fact that knowledge is more accepted, established and plausible for future usage in the scientific community. Specifically, we emphasize on the number of people worked on finding three factors on the knowledge source, 1) number of people worked on the knowledge curation and compilation, 2) number of people collecting the knowledge and the knowledge is still (partially/fully) unexplored, 3) number of people validating and verifying the knowledge. This classification aids us in explaining our proposal in this paper, as well as extend any other additional sources of causal knowledge fit into this proposition of hierarchy.

\begin{table}[ht]
\small
\begin{tabular}{|p{1.8cm}|p{3cm}||p{2.3cm}|p{2.3cm}|p{2.3cm}||p{2cm}|}
\hline
\textbf{Knowledge Source} & \textbf{Example} & \textbf{Curated Knowledge collected from (no. of individual)} & \textbf{Uncurated/ unexplored Knowledge from (no. of individual)} & \textbf{Verified by (no. of individuals)} & \textbf{Tier (Weight)} \\
\hline
Background Knowledge & Friend applying cold compress to reduce swelling & 1 & N/A & 0 & $CK_E (W_E)$ \\
\hline
Expertise & Physician prescribing Vit D supplement to improve fatigue & 1 & N/A & 1 & $CK_E (W_E)$ \\
\hline
General Article/ Opinion  & Newspaper article on related topic & 1 & N/A & 1 / \textgreater 1 (many) & $CK_E (W_E)$\\
\hline
Data & EHR Dataset & N/A & \textgreater 1 (many) & \textgreater 1 (many) & $CK_D (W_D > W_E)$ \\
\hline
Peer-reviewed Literature  & Published paper on Aspirin efficacy on treating headache  & \textgreater 1 (many) & \textgreater 1 (many) & \textgreater 1 (many) & $CK_L (W_L > W_D)$ \\
\hline
\end{tabular}
\caption{High-level classification of causal knowledge sources for CKH}
\label{tab:ckh_basic}
\end{table}


\subsection{Algorithm}

Our algorithm works with the following inputs: problem statement ($\mathbf{PS}$), defined keywords ($\mathbf{K}$), (empty) structural causal model ($\mathbf{SCM}$), pre-defined tier weights ($\mathbf{W}$), and inter-rater agreement function ($\mathbf{IRR}$). For each of the three tiers of CKH, we encode specific and relevant causal knowledge within the tier, and systematically update the knowledge base to derive the causal structure. 

We start with an empty SCM, with no values assigned to $\mathbf{U}$, $\mathbf{V}$, $\mathbf{F}$ and $\mathbf{P(u)}$ (\textit{introduced and discussed in the supplementary document}). For each individual tier of causal knowledge hierarchy, we follow the steps described below:
\begin{enumerate}
    \item \textbf{Encode tier-specific information:} For each tier, we encode all the information in the causal graph specific to that tier. Specifically, we encode: 
    \begin{enumerate}
        \item experts' opinion and background knowledge as individual edges, and their confidence score (between $0$ and $1.0$) in those edges (with directions), in Tier 1 ($\mathbf{CK_E}$),
        \item edges generated by the causal structure learning algorithms run on the data sets, in Tier 2 ($\mathbf{CK_D}$), and,
        \item causal relationships extracted from literature as directional edges in Tier 3 ($\mathbf{CK_L}$).
    \end{enumerate} 
    
    \item \textbf{Develop a scoring matrix from encoded information:} From the information encoded for each tier, we build a causal information-based scoring matrix $P$ for the specific tier, with a dimension of $m \times n$. Here $m$ is the number of rows equal to the count of unique pairs of nodes (variables). For $q$ number of total variables ($U \cup V$) in an individual tier, we have $m = {\binom{q}{2}} = \frac{q(q-1)}{2}$. $n$ is the number of columns in the scoring matrix $P$. For a specific row in $P$ with the nodes (or variables) $A$ and $B$, we have four columns signifying the type of causal connection between them: i) $A \rightarrow B$, ii) $A \leftarrow B$, iii) no causal connection between $A$ and $B$ and, iv) no causal information available between $A$ and $B$. The complete matrix represents the causal knowledge in the specific tier. 
    
    \item \textbf{Compute individual edge confidence based on agreement from scoring matrix}: Next we calculate individual edge confidence from each row in the scoring matrix $P$, through plurality voting. For each pair of nodes $A, B$ in $P$, we iterate through rows $i$ of $P$ and use the equation of edge confidence of causal connections between variables $A$ and $B$:

    \begin{equation}
        q_i = [ P_i(n) ] \textrm{ for } n = 0,1,2\\
    \end{equation}
    \vspace{-3mm}
    \begin{equation}
        e_{A,B} = \frac{max(q_i)}{\sum q_i}
    \end{equation}
    
    Here, $e_{A,B}$ signifies the confidence of the causal connection or on the directional edge between variables $A$ \& $B$, $q_i$ represents the first three values of row $i$ of scoring matrix $P$ with variables $A$ and $B$.
    
    \item \textbf{Estimate agreement score from the scoring matrix:} Using the inter-rater agreement function ($IRR$), we calculate the agreement score ($\mathbf{\alpha}$) (i.e., Fleiss' kappa, varies from 0 to 1) from the generated scoring matrix $P$ (\textit{explained further under background in supplementary document}).
    
    \item \textbf{Compute individual weighted edge confidence:}
    Next, we calculate the weighted edge strength for all edges within a tier, using the equation:
    \begin{equation}
        weighted\_e_{A,B} = e_{A,B} \times \alpha \times W_i
    \end{equation}
    Here, $weighted\_e_{A,B}$ signifies the weighted edge confidence of the causal connection between variables $A$ \& $B$, $\alpha$ is the agreement score calculated previously, and $W_i$ represents the weight of the specific tier ($W_E$ / $W_D$ / $W_L$). Within a specific tier, only $e_{A,B}$ is different for individual edges, whereas $\alpha$ and $W_i$ remains the same. In the best case, where edge weight for a specific edge and agreement score are both $1.0$, the weighted edge strength can be the maximum (the weight of the Tier).
    
    \item \textbf{Extract tier-specific insights:} 
    From the generated weighted edge confidences $weighted\_e_{A,B}$, we extract tier-specific causal insights and carry them forward to the next tier. Specifically, in Tier 1 ($\mathbf{CK_E}$), we set a predefined confidence threshold to select edge whitelist. Any edge with edge confidence over the threshold is put in the whitelist and used in the structure learning algorithms in Tier 2 ($\mathbf{CK_D}$). Similarly, in Tier 2 ($\mathbf{CK_D}$), we extract an incomplete causal structure skeleton $\langle U, V, F \rangle$ and probability distribution $\langle P(u) \rangle$, to carry over to the next tier. Finally, in Tier 3 ($\mathbf{CK_L}$), we extract the complete SCM $\langle U, V, F, P(u) \rangle$, with encoded information from all tiers of the CKH.
    
    
    \item \textbf{Move extracted insights to next step:}
    Next we get the output of the specific tier, the extracted insights and use them in the next tier as inputs.
\end{enumerate}

\paragraph{Edge Orientation Step:}
In this step, we check for any potential cycles between variables, and re-orient them prioritizing the weighted edge strengths, and generate a complete directed acyclic graph. For this task, we implement the edge orientation process from PC algorithm \citep{scutari2019learns}. For each triplet of nodes, $A-B-C$, we recursively set edge directions using the two rules: (a) if $A$ is adjacent to $B$ and there is a strictly directed path from $A$ to $B$, we then replace $A-B$ with $A\rightarrow B$ (to avoid introducing cycle), and (b) if $A$ and $C$ are not adjacent and we have $A\rightarrow B$ \& $B-C$, we replace $B-C$ with $B\rightarrow C$ (to avoid introducing new v-structures). The complete directed acyclic graph, along with the computed $\langle U, V, F, P(u) \rangle$ is the resultant SCM for the problem domain.
The algorithmic pseudocode is presented in \ref{alg:main}.








%% file: sections/algorithm.tex
\begin{algorithm2e}[htbp]
\caption{Structural Causal Model Estimation using Causal Knowledge Hierarchy (CKH)}
\label{alg:main}
\KwIn{$PS$, $K$, $SCM=NULL$, $W=[W_E, W_D, W_L]$, $IRR$}
\KwOut{$SCM_m=\langle U,V,F,P(u),CONF \rangle$}
Initialize empty confidence for all edges: $CONF \gets \phi$ \;
Update modified $SCM$ for output: $SCM_m \gets SCM + CONF$ \;
\tcc{(a) Tier 1: Causal Knowledge from Experts' opinion ($CK_E$):}
Select group of experts: $Exp \gets [exp_1, exp_2, exp_3, ...]$\;
\While{expert $exp_i$ in $Exp$}{
    $CR'_{CK_E}[i] \gets extract\_causal\_relationships(exp_i,PS, K, U \cup V = \phi)$\;
}
$U, V \gets get\_vars(CR'_{CK_E})$\;
\While{expert $exp_i$ in $Exp$}{
    $CR''_{CK_E}[i] \gets extract\_causal\_relationships(exp_i,PS, K, U \cup V)$\;
}
$U, V \gets get\_vars(CR'_{CK_E} \cup CR''_{CK_E})$\;
$P \gets create\_grading\_tuple(CR'_{CK_E} \cup CR''_{CK_E})$\;
$F, CONF \gets compute\_weighted\_confidence(W_E,P, IRR, SCM_m)$\;
$SCM_m \gets update\_scm(SCM_m, [U, V, F, CONF])$\;
\tcc{(b) Tier 2: Causal Knowledge from Data ($CK_D$):}
Using $U \cup V$ and $K$, gather relelvant datasets: $D = [d_1, d_2, d_3,...]$\;
Select different causal structure learning algorithms: $SLA = [sla_1, sla_2, sla_3]$\;
\While{output $model$ in $SLA \times D$}{
    $CR'_{CK_D}[i] \gets extract\_causal\_relationships(model)$\;
}
$U, V \gets get\_vars(CR'_{CK_D})$\;
$P(u) \gets get\_probability\_distribution(D)$\;
$P \gets create\_grading\_tuple(CR'_{CK_D})$\;
$F, CONF \gets compute\_weighted\_confidence(W_D, P, IRR, SCM_m)$\;
$SCM_m \gets update\_scm(SCM_m, [U, V, F, P(u), CONF])$\;

\tcc{(c) Tier 3: Causal Knowledge from Literature ($CK_L$):}
Using $PS$ and $K$, gather relevant literature: $L = [l_1, l_2, l_3,...]$\;
$U, V \gets get\_vars(L)$\;
\While{literature $l$ in $L$}{
    $CR'_{CK_L}[i] \gets extract\_causal\_relationships(L)$\;
}
$P \gets create\_grading\_tuple(CR'_{CK_L})$\;
$F, CONF \gets compute\_weighted\_confidence(W_L,P, IRR, SCM_m)$\;
$SCM_m \gets update\_scm(SCM_m, [U, V, F, CONF])$\;
\tcc{(d) Edge orientation:}
$SCM_m \gets orient\_edges(SCM_m)$\;
\KwRet{$SCM_m$}
\end{algorithm2e}

%% file: sections/4_simulation.tex
\section{Experimental Results}
For our experiments, we rely on \emph{a simulation with a standard causal model}. We use pre-defined default values for hyper-parameters, and validate our results with the initial ground truth causal model and provide sensitivity analysis of the SCM.

\begin{figure}[ht]
    \centering
    \includegraphics[width=\linewidth]{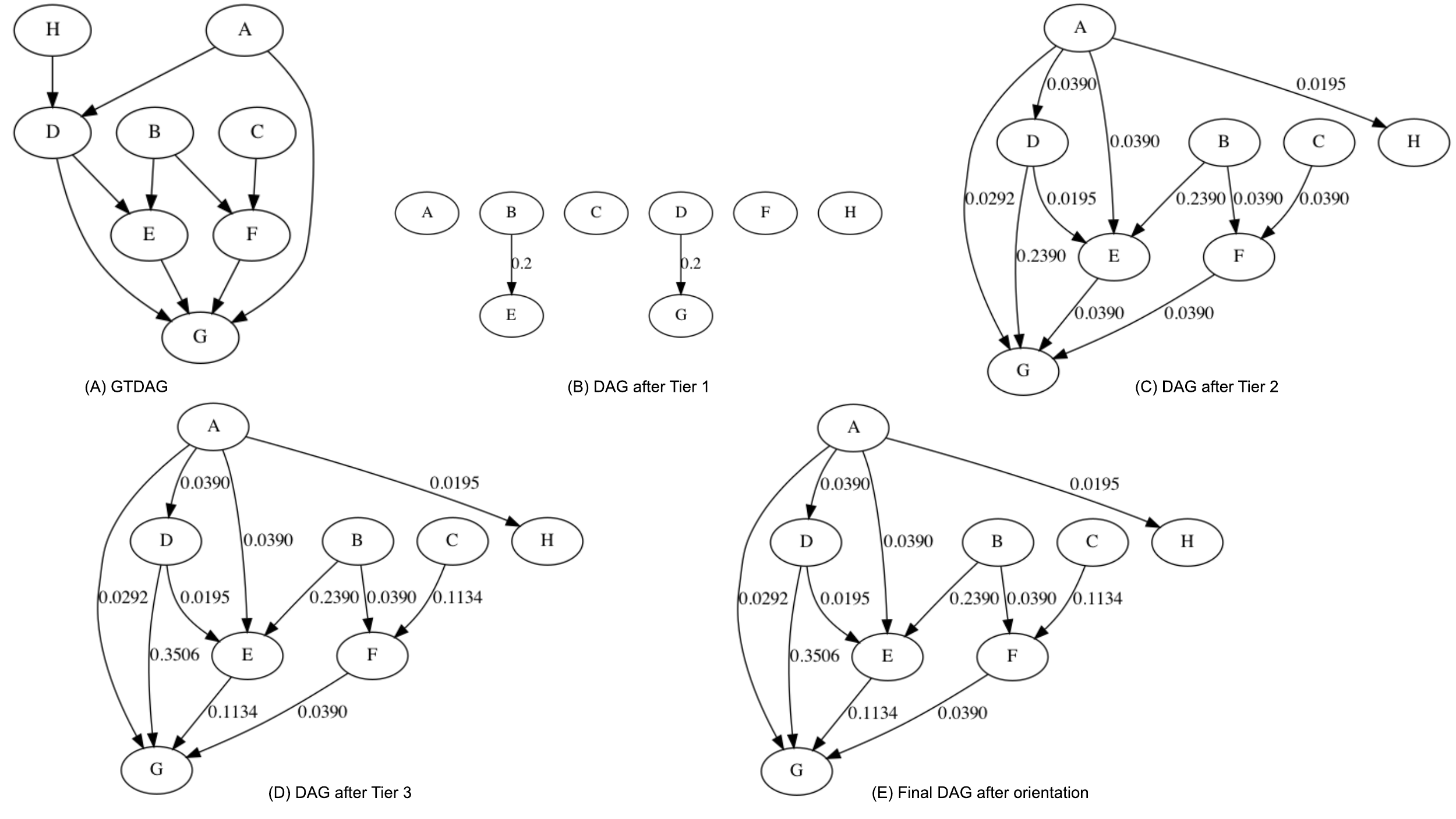}
    \caption{Ground-truth Causal Model versus Structural Causal Models with edge confidences at individual tiers of CKH}~\label{fig:sim1}
\end{figure}

\paragraph{Ground Truth Causal Model}
We start the simulation with a causal model with a ground truth SCM (\autoref{fig:sim1}), (a). For this, we refer to the ``clgaussian" dataset from `bnlearn' library \citep{bnlearn-clgaussian}. The dataset has 5000 data-points and is generated from a causal model with one normal (Gaussian) variable, four discrete variables and three conditional Gaussian variables. For validation, we assume this initially defined causal model to be the Ground Truth Directed Acyclic Graph (GTDAG). 

\paragraph{Optimization Function}
 Keywords are defined with a complete set of variables $\{A, B, C, D, E, F, G, H\}$. We define the optimization problem \emph{to identify the causal effect and relations between variables $D$ and $G$}, along with all associated variables $\{A, B, C, E, F\}$ with the best-fitting GTDAG for the SCM. We also hypothesize that the domain is well-explored \textit{(with sufficient experts, data and literature)} and set the values of weights as $W_{T1}=0.2, W_{T2}=0.3, W_{T3}=0.5$, as well as confidence threshold for Tier 1 $threshold = 0.8$.

\paragraph{Tier 1: Causal Knowledge from Experts' Opinion}
For Tier 1, we consider only one expert $Exp=\{exp_1\}$, and use their knowledge, expressed through causal graphs \textit{(although in reality, we might have multiple experts)}. We go through this step in two phases, the first causal graphs are encoded from each of the experts by describing the problem statements ($PS$) and keywords ($K$). We integrate all the variables and create a super-set of nodes, $U \cup V$. The second causal graph is derived after providing $U \cup V$, along with both $PS$ and $K$ to the experts. This generates $2 x$ causal graphs from experts' opinions based on the problem statement. From these graphs, we generate simplified causal connections:

\begin{itemize}[noitemsep,nolistsep]
    \item $A \rightarrow D$, confidence: 0.5
    \item $D \rightarrow G$, confidence: 1.0
    \item $B \rightarrow E$, confidence: 1.0
\end{itemize}

Based on the pre-defined confidence threshold for Tier 1, we derive a combined causal graph from them. In the combined causal graph in Tier 1, we have $M = < U, V, F, P(u)=\phi, CONF=\phi>$, where $U$ and $V$ comes from the super-set of variables suggested by the experts, and $F$ is defined from the causal relationships suggested by the experts. We calculate the agreement score (using Kappam' fleiss), and measure total confidence values using the agreement score and $W_{T1}=0.2$. Since we use one expert in this experiment, the agreement score for this tier is $1.0$, and we use this value in Equation 3 to compute the weighted edge confidence. We build our scoring matrix $P$ based on the weighted edge confidence higher than the confidence threshold. The resulting causal graph after Tier 1 is presented in \autoref{fig:sim1}, (b).

\paragraph{Tier 2: Causal Knowledge from Data}
For Tier 2, we generate three separate datasets, $D=\{d_1, d_2, d_3\}$ based on the GTDAG. To simulate a varying number of datasets in real-world, we generate and use three different datasets for which the underlying causal relations are invariant. $d_1$, is generated from variables $\{A, D, G\}$, $d_2$, is generated from only variables $\{B, D, E, G, H\}$, and $d_3$, is generated from all the variables.  We use two specific structure learning algorithms, $SLA=\{sla_1, sla_2\}$, with $sla_1$ being \textbf{PC algorithm} and $sla_2$ being \textbf{MMHC algorithm}. Additionally, we use edge whitelists from Tier 1 for structure learning. We show outputs of all SLAs on datasets in the supplementary document, Figure 3. 

We update the scoring matrix $P$ for this tier and compute weighted edge confidence. We resolve contradictory edges between Tier 1 and Tier 2 by selecting the highest weighted edge confidence of the two. We sum up the weighted edge confidences and the resulting causal graph after Tier 2 is shown in \autoref{fig:sim1}, (c).

\paragraph{Tier 3: Causal Knowledge from Literature}

In Tier 3, we go through peer-reviewed literature for the problem domain. For this simulation, we assume three causal information sets, $L=\{l1, l2, l3\}$, each of which is extracted from individually published literature. The summary of causal relationships extracted from the literature $L$ is:

\begin{enumerate}[noitemsep, nolistsep]
    \item Literature 1 ($l1$): $D \rightarrow G$, $A \rightarrow D$
    \item Literature 2 ($l2$): $F \rightarrow G$, $E \rightarrow G$, $D \rightarrow G$, $C \rightarrow F$
    \item Literature 3 ($l3$): $B \rightarrow E$, $C \rightarrow F$, $D \rightarrow E$, $D \rightarrow G$, $E \rightarrow G$
\end{enumerate}

Similar to the previous tiers, we update the scoring matrix $P$ and compute weighted edge confidence as well. We also resolve conflicting edges, \textit{(if any)}, between new causal graph in Tier 3 and original causal graph after Tier 2, depending on edge confidence. Finally, we sum up the weighted edge confidences with previous tiers and the resulting causal graph after Tier 3 is shown in \autoref{fig:sim1}, (d). A summary of the edges with edge orientations and combined weighted edge confidence is shown in the supplementary document, Table 3. 

\paragraph{Edge Orientation}

In the last stage of edge orientation, we see whether any cycles were created in the process. In case one is found, we follow edge orientation process \textit{(as described in method section)} and derive the updated SCM. We present the eventual output in \autoref{fig:sim1}, (e).

\begin{table*}[ht]
    \centering
    \caption{Iterations of simulations with incorrect edge directions injected for each tier (one edge incorrect (A1), two edges incorrect (A1+A2), and three edges incorrect (A1+A2+A3)). Corresponding agreement scores and performance metrics (TPR: True Positive Rate, FDR: False Discovery Rate, MCC: Matthews Correlation Coefficient) are reported.}
    \label{tab:result-sensitivity}
  \begin{tabular}{clcccccc}
    \toprule
    && \multicolumn{3}{c}{\textbf{Agreement scores}} & \multicolumn{3}{c}{\textbf{Metrics}} \\
    \cmidrule(lr){3-5} \cmidrule(lr){6-8} 
    \multicolumn{2}{c}{\textbf{Change in edge}} & \textbf{Tier 1} & \textbf{Tier 2} & \textbf{Tier 3} & \textbf{TPR} & \textbf{FDR} & \textbf{MCC} \\
    \midrule
    \multicolumn{2}{c}{No alteration} & 1.0 & 0.13 & 0.372 & 0.8929 & 0.1320 & 0.7846 \\
    \midrule
    \multirow{3}{*}{Tier 1} & A1 & 1.0 & 0.13 & 0.372 & 0.8571 & 0.1563 & 0.7244 \\
    & A1+A2 & 1.0 & 0.13 & 0.372 & 0.8214 & 0.1636 & 0.6535 \\
    & A1+A2+A3 & 1.0 & 0.13 & 0.372 & 0.8214 & 0.1636 & 0.6535 \\
    \midrule
    \multirow{3}{*}{Tier 2} & A1 & 1.0 & 0.124 & 0.28 & 0.8929 & 0.1320 & 0.7846 \\
    & A1+A2 & 1.0 & 0.115 & 0.28 & 0.8929 & 0.1320 & 0.7846 \\
    & A1+A2+A3 & 1.0 & 0.107 & 0.28 & 0.8929 & 0.1320 & 0.7846 \\
    \midrule
    \multirow{3}{*}{Tier 3} & A1 & 1.0 & 0.13 & 0.319 & 0.8929 & 0.1320 & 0.7846 \\
    & A1+A2 & 1.0 & 0.13 & 0.28 & 0.8929 & 0.1320 & 0.7846 \\
    & A1+A2+A3 & 1.0 & 0.13 & 0.28 & 0.8929 & 0.1320 & 0.7846 \\
    \bottomrule
  \end{tabular}
\end{table*}

\paragraph{Evaluation}

For evaluating our proposed method, we compare the output SCM of the algorithm with the GTDAG. Specifically, we compare edges with directions from our proposed method with that of the GTDAG, and report the true positive rate (TPR), false discovery rate (FDR) and Matthews Correlation Coefficient (MCC). An edge-by-edge comparison of generated output DAG with that of GTDAG is considered as a classification problem \citep{tsamardinos2006max,scutari2019learns}, and the reported performance metrics are frequently reported for similar causal graph identification problem \citep{guo2020survey}. For the simulation, with a node number of $q=8$, we check for $\binom{n}{2} = 28$ edges' causal directions. On average, our proposed method achieved a TPR of $\mathbf{89.29\%}$ \textit{(the closer to $\mathit{100\%}$ TPR is, the better)}, FDR of $\mathbf{13.20\%}$ \textit{(the closer to $\mathit{0\%}$ FDR is, the better)}, and MCC of $\mathbf{0.7846}$ \textit{(the closer to $\mathit{+1}$ MCC is, the better)}.

It is possible to incorporate incorrect causal knowledge due to biased opinion, dataset, or publication. For this, we additionally perform sensitivity analysis with perturbed edges  within individual tiers. We randomly select three edges and alter their directions. Specifically, we add the following information:

\begin{table*}[ht]
    \centering
    \caption{Alteration in causal edges in the simulation}
    \label{tab:my_label}
    \begin{tabular}{cccc}
        \toprule
         \textbf{Id} & \textbf{Change in simulation} & \textbf{Ground truth} & \textbf{Altered information} \\
         \midrule
         A1 & Add false causal edge & $C \dots D$ & $C \rightarrow D$ \\
         A2 & Alter true causal edge & $E \rightarrow G$ & $E \leftarrow G$ \\
         A3 & Remove true causal edge & $B \rightarrow F$ & $B \dots F$ \\
         \bottomrule
    \end{tabular}
\end{table*}



We run the simulation initially without any alteration, and then with multiple perturbed edge directions. For each tier, we perturb one edge (A1), two edges (A1+A2), and three edges (A1+A2+A3), and report the general accuracy, precision, recall, and F1-score, along with the change in agreement scores in each case. \autoref{tab:result-sensitivity} shows the reported outcomes for each simulations.

In general, with gradual perturbation, performance metrics as well as agreement scores decrease, however this decrease is not drastic, due to the weights of tiers of CKH. For each tier, agreement scores decrease, however the decrease in agreement score does not necessarily alter the outcome. 

%% file: sections/5_discussion.tex
\section{Discussion}

Identifying causality is a critical part of many analyses, specifically in clinical research where trust in models is low, and safety and efficacy of clinical decisions is essential. In that context, SCM provides the theoretical foundation for identifying causation from large datasets. However, the lack of methodologies to derive an SCM for estimating causal effect is a fundamental research problem. We have proposed a novel methodology to combine causal knowledge from various sources such as experts' opinion, data, and literature to derive domain-specific accurate SCMs. We discuss the importance of causal information from sources other than just data, and present the rationale behind using hierarchy of causal knowledge. As demonstrated by our experiments, our proposed method (CKH) effectively identifies the most compatible causal models, with higher accuracy and F1-score, from opinions of experts working in the field, outputs of SLAs on existing data, and reported information in peer-reviewed publications. Further discussion is addressed in the appendix.

The CKH-driven algorithm relies on availability and abundance of causal knowledge sources, making it unreliable when there is a lack of causal knowledge from multiple tiers of sources (e.g., CKH-generated causal model from only data will have low confidence compared to that generated from all three sources). Similar to other open research problems in causal inference, it cannot verify a ground-truth SCM. An alternate to CKH is to not make strong inferences about causal directions to build a DAG and rather derive a Markov equivalence class. Within individual tiers under CKH for a specific problem domain, there is a challenge in finding and selecting experts. Similar difficulty arises within Tier 3 (literature, $CK_L$), since extraction of causal knowledge from literature is itself a research problem under Natural Language Processing (NLP) \citep{wood2018challenges,keith2020text} and is currently being tackled by NLP researchers.

%% file: sections/6_acknowledgments.tex
\acks{We thank the collaborators and the research lab members for their valuable input (
\textit{anonymous}). The icons of the images were created by various artists (Vectors Market, Linector, Becris, Freepik) from Flaticon (flaticon.com).}

%% file: sections/8_appendix.tex
\clearpage
\appendix

\section{Background}

Before presenting our proposed methodology we briefly discuss the relevant scientific concepts needed to explain the individual steps as well as the rationale behind them. The objective of our method is to generate an expert augmented \textbf{Structural Causal Model (SCM)} for a specific hypothesis, where part of the causal information used to generate the graphical representation comes from the application of \textbf{Structure learning algorithms} on datasets. The building blocks of our proposed methodology includes structural causal models (SCM), \textbf{Inter-rater agreement functions} for generating aggregated information and a Causal Knowledge Hierarchy inspired from \textbf{Hierarchy of Evidence} in evidence-based health research. These concepts are introduced in this section with details. 

\subsection{Structural Causal Models}

Developed on the foundations of probabilistic graphical models, SCMs are graphical representations of the causal relationships between variables, and are used to draw causal inferences. An SCM is often expressed by a causal graph $G$. Each node $V$ in $G$ represents an observed or unobserved variable, and each directed edge $E$ represents the causal relationships between them. 

An SCM $M$ is a 4-tuple $\langle U, V, f, P(u) \rangle$ \citep{pearl2009causality} where,

\begin{itemize}
    \item $U$ is a set of background (exogenous) variables that are determined by factors outside of the model, 
    \item $V$ is a set of observable (endogenous) variables that are determined by variables in the model, 
    \item $F$ is a set of functions such that each $f_i \subseteq F$ is a mapping from the respective domains of $U_i \cup P A_i$ to $V_i$, where $U_i \subseteq U$ and 
    \item $P A_i \subseteq V \setminus V_i$ and the entire set $F$ forms a mapping from $U$ to $V$, and $P(u)$ is a probability distribution over the exogenous variables.
\end{itemize}

\begin{figure}[htbp]
    \centering
    \includegraphics[width=0.4\textwidth,trim=0mm 42mm 0mm 74mm, clip]{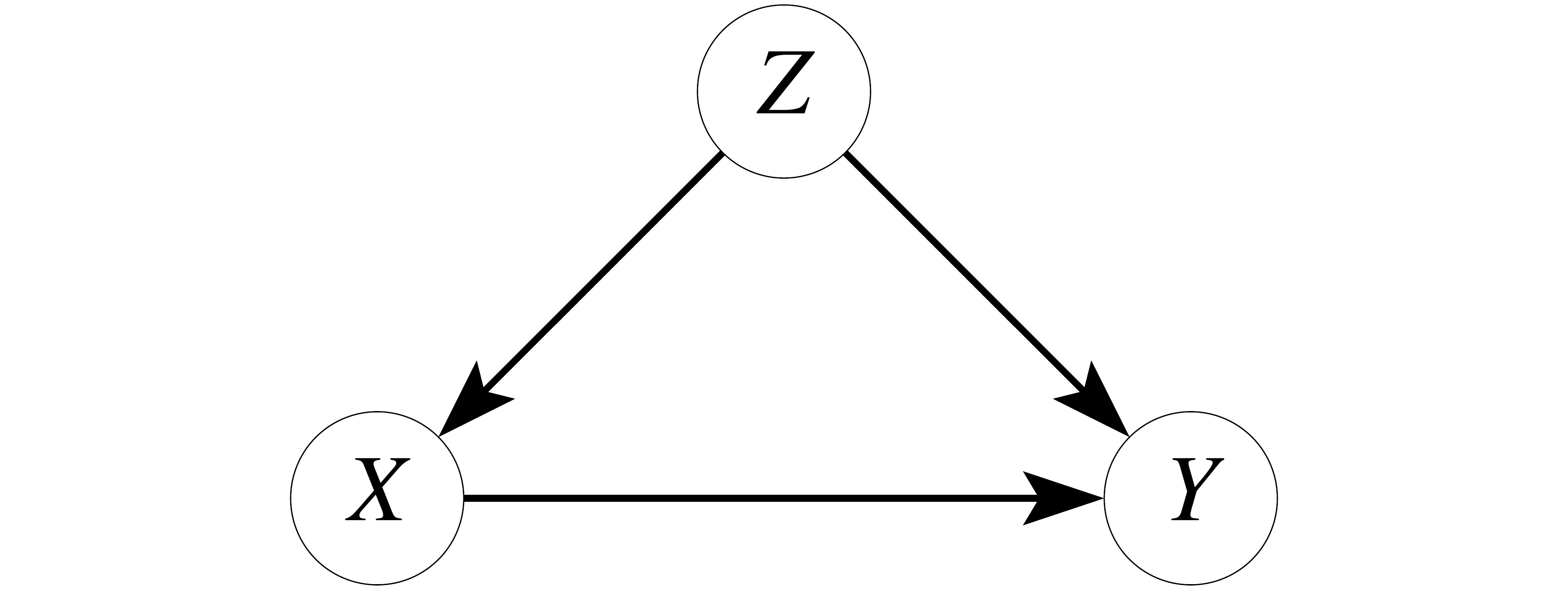}
    \caption{A simple graphical model representing observational study with three variables, with treatment $X$, outcome $Y$ and confounder $Z$, expressed using causal directed acyclic graph \textit{(nodes are the variables, edges portray causal relationships between variables)}}
    \label{fig:simple_dag}
\end{figure}

To find the causal effect of variable $X$ on variable $Y$, do-calculus is introduced \citep{pearl2016causal}. Do-calculus comes with its own set of strong mathematical tools, such as, rules of do-calculus, backdoor criterion, that is used to map the observational reality to the corresponding experimental reality with the identifiability equation by adjusting for different kinds of biases (e.g., confounding bias), if it exists. 

\figureref{fig:simple_dag} represents a simple graphical model for an SCM. Although an SCM is used to represent the underlying causal model, in reality, the ground truth causal model in social sciences or medical sciences is never fully known \citep{freedman1999there}. The causal graph usually represents a set of assumptions explicitly in the problem domain of interest. Given a data set, an SCM can be any of the causal model for a Markov equivalence class \citep{spirtes2010introduction}, meaning multiple causal models can be true within a Markov equivalence class for a given data set. Consequently, the validation of a causal model is one of the fundamental challenges in causal inference research. The state of the art generates the most-fitting SCM from datasets using structure learning algorithms \textit{(described in next subsection)} using the properties of conditional probability distributions.

\subsection{Structure Learning Algorithms}

Other than domain expertise, observational or experimental data can be used to generate a causal graph. Data are the result or snapshot of the underlying causal mechanisms between variables. To recover the causal relationships from data, a rich set of algorithms have been developed over the past thirty years  \citep{spirtes2000constructing,tsamardinos2006max,shimizu2014lingam}. Causal structured learning is where we try to learn the causal graph or aspects of the causal mechanism. The problem is fundamentally a model selection problem, and these algorithms are called structured learning algorithms (SLA) \citep{heinze2018causal,drton2017structure}, where a graph is learned or estimated that best describes the dependence structure in a given data set. The learning process includes relying on necessary assumptions (i.e., causal sufficiency, causal faithfulness, linearity), finding conditional dependencies between variables (i.e., Bayes' theorem) and differentiating between different causal structures (i.e., chains, forks, colliders). 

Specifically, learning an SCM (or, Bayesian Networks) with a directed acyclic graph (DAG) $G$ and parameters $\theta$ from a dataset $D$ with $n$ observations is completed in two steps \citep{scutari2019learns}: (1) finding the DAG $G$ which encodes the dependence structure of data $D$, called structured learning, and (2) estimating the parameters $\theta$, given the obtained $G$ from structured learning, called parameter learning: $$P(G,\theta | D) = P(G | D) \cdot P(\theta|G,D)$$ 
Consequently, SLAs are a key component in estimating causal effects within a dataset.

Several algorithms have been proposed in the literature for SLAs \citep{heinze2018causal}, however they differ in their approaches, assumptions, and graphical objects generated. This makes their outcomes varying (even based on the same data source) and difficult to compare. 

The main classes of existing SLAs \citep{heinze2018causal} are:
\begin{enumerate}
    \item \textbf{Constraint-based methods:} Peter-Clark (PC), rankPC , fast causal inference (FCI), and rankFCI
    \item \textbf{Score-based methods:} greedy equivalence search (GES), rankGES, greedy interventional equivalence search (GIES), and rankGIES
    \item \textbf{Hybrid methods:} Max-min hill climbing (MMHC)
    \item \textbf{Structural equation models with additional restrictions:} linear non-Gaussian acyclic models(LINGAM)
\end{enumerate}

An overview of their generated graphical models and assumptions required are summarised in \tableref{tab:table_sla}. 

\begin{table*}[htbp]
  \floatconts
  {tab:table_sla}
  {\caption{Summary of different SLAs and their outputs}}
  {\begin{tabular}{lllllll}
    \toprule
     & PC & FCI & GES & GIES & MMHC & LINGAM \\
    \midrule
    Causal sufficiency & Yes & \textit{No} & Yes & Yes & Yes & Yes \\
    Causal faithfulness & Yes & Yes & Yes & Yes & Yes & \textit{No} \\
    Acyclicity & Yes & Yes & Yes & Yes & Yes & Yes \\
    Non-gaussian errors & \textit{No} & \textit{No} & \textit{No} & \textit{No} & \textit{No} & Yes  \\
    Known do-intervention & \textit{No} & \textit{No} & \textit{No} & Yes & \textit{No} & Yes  \\
    Output & CPDAG & PAG & CPDAG & PDAG & DAG & DAG  \\
    \bottomrule
  \end{tabular}}
\end{table*}

Since different SLAs can generate different SCMs from the same datasets, there is a need for a principled approach for combining information, which is also correlated with the agreement between them. For this purpose, we leverage inter-rater agreement functions \textit{(described in next section)} to generate an aggregated graphical model that best represents the data along with other sources of causal information (e.g., output SCMs of SLAs, expert opinion or peer-reviewed literature).

\subsection{Inter-rater Agreement}
Inter-rater agreement \citep{mchugh2012interrater} is the degree of agreement among raters, which generates a score on homogeneity, or consensus, in the ratings given by judges or raters. In causal inference we frequently arrive at multiple ratings on causal relationships (by experts' opinion, or from outputs of SLAs), and this is a mechanism to mitigate the discrepancy. To the best of our knowledge, this mechanism has not been previously used in the context of causal graph generation. 

Inter-rater agreement function relies on three operational definitions of agreement \citep{saal1980rating}:

\begin{enumerate}[noitemsep,nolistsep]
    \item Reliable raters agree with the ``official" rating of a performance.
    \item Reliable raters agree with each other about the exact ratings to be awarded.
    \item Reliable raters agree about which performance is better and which is worse.
\end{enumerate}

In addition, reliable raters are assumed to behave as independent witnesses to the model where they express their independence by disagreeing slightly. In our proposed methodology, we assume the expert opinion, literature, or SLAs are independent raters of causal relationships who capture and express their judgements based on their individual knowledge sources.

Different types of inter-rater agreement functions and scores have been proposed, each with their unique features and strengths. We present a brief overview \citep{mchugh2012interrater} of a few of them here. 

\begin{enumerate}[noitemsep,nolistsep]
    \item Percent agreement 
    \item Cohen's kappa coefficient
    \item Fleiss kappa \footnote{Adaptation of Cohen’s kappa for 3 or more raters}
    \item Joint probability of agreement
    \item Pearson r coefficient
    \item Krippendorff’s alpha \footnote{Useful when there are multiple raters and multiple possible ratings}
\end{enumerate}

Along with inter-rater agreement function applied on rated causal relationships between variables by raters or algorithms, we propose to incorporate the well established \textbf{Hierarchy of Evidence} in evidence-based health research in our methodology.

\subsection{Hierarchy of Evidence}
A hierarchy of evidence is needed when there are multiple results or inferences from similar scientific studies (sometimes even contradictory) and one has to choose one or combine them. Hierarchy of evidence (or, levels of evidence) is a scoring that quantifies the rank or strength of the results or outcomes from scientific and experimental studies. The hierarchy relies on the study design, validity and applicability to patient care, and quality of data \citep{ackley2008evidence}. When choosing between multiple findings from experimental studies, hierarchy of evidence is critically important. For example, in healthcare professionals are required to decide on clinical actions based on the best evidence available. One of the most significant reasons behind using a hierarchy of evidence is to upgrade quality of care, by identifying and promoting practice that is effective and by eliminating those who are ineffective or harmful \citep{akobeng2005principles}.

Different hierarchy of evidence have been proposed in the literature, based on design of studies and the endpoints measured. A commonly accepted level of effectiveness rating scheme \citep{ackley2008evidence} is presented in \figureref{fig:ebm-pyramid}.


\begin{figure}[htbp]
    \floatconts
    {fig:ebm-pyramid}
    {\caption{Evidence-Based Medicine (EBM) Resources}}
    {\includegraphics[width=0.9\linewidth]{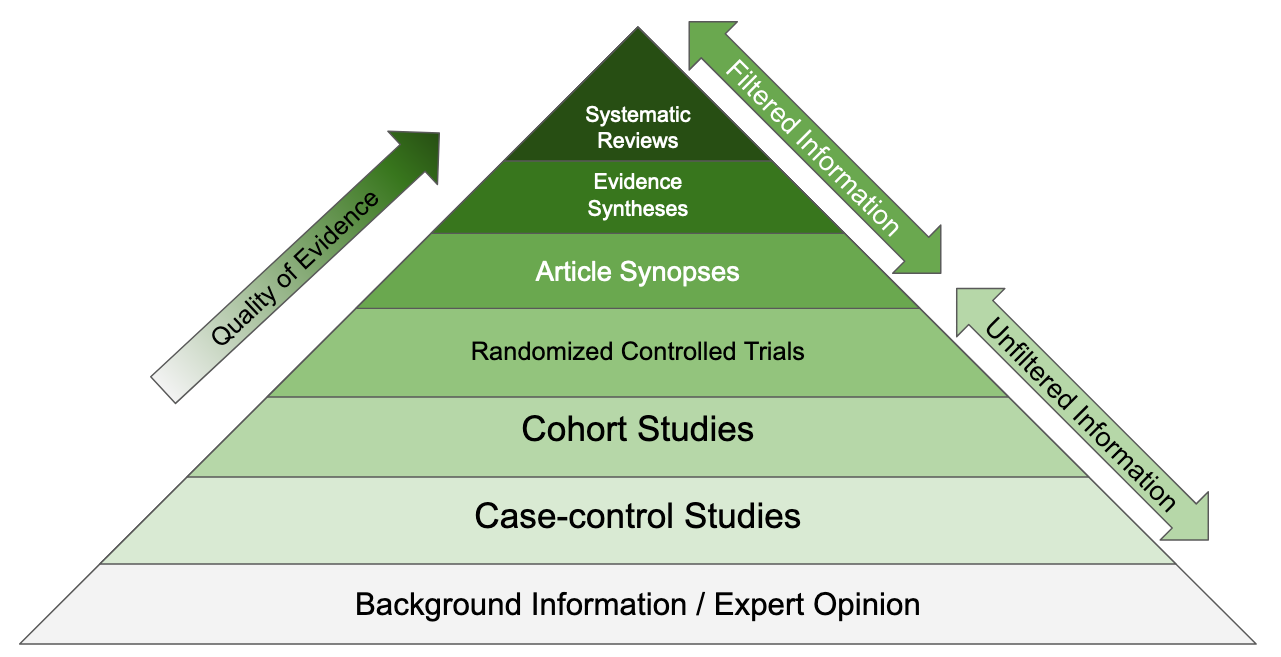}}
\end{figure}

\section{Method}

The following table discusses a potential dictionary for translating causal knowledge from literature to the syntax of structural theory of causation. The table is not comprehensive and part of an ongoing solution to this research problem.

\begin{table*}[htbp]
    \centering
    \caption{Causal knowledge ($CK_L$) extraction from peer-reviewed literature}~\label{tab:pub}
    \begin{tabular}{llll}
        \toprule 
        \textbf{Section in} & \textbf{Information} & \textbf{Translation to} & \textbf{do-Calculus} \\
        \textbf{Publication} & & \textbf{Causal Discovery} & \\
        \midrule  
        Objective / & Objective & Causal Query & $P(y|do(x))$ \\
        Outcome measures & & & \\
        \midrule 
        Objective / & Effect of \_\_ & Treatment & $X$ \\
        Interventions & & & \\
        \midrule 
        Objective / & Effect on \_\_ & Outcome & $Y$ \\
        Outcome measures & & & \\
        \midrule 
        Methods & Patient & Covariates & $Z$\\
        & demographic & (potential confounders) & \\
        \midrule 
        Methods & Study type & Backdoor from & $Z \rightarrow X$ \\
        & & confounders to treatment & \\
        \midrule 
        Methods /  & Inclusion \&  & Target group & $S=1$ \\
        Results & Exclusion criteria & & \\
        \midrule 
        Results / & Findings & causal effect & Value of \\
        Conclusion & & & $P(y|do(x))$ \\
        \bottomrule 
    \end{tabular}
\end{table*}


\section{Algorithm}

The following algorithmic pseudocode discusses the process of computing weighted edge confidence for specific edge within a tier. The pseudocode is a part of the complete algorithm discussed in the paper.

\begin{algorithm2e}[htb]
\caption{Computation of Weighted Confidence for Individual Tier}
\label{alg:weight}
\KwIn{$W, P, IRR, SCM$}
\KwOut{$F, CONF$}
    Measure agreement score: $\alpha \gets IRR(P)$\;
    $CONF \gets \phi$\;
    \For{each edge in $SCM$}{
        Equation 1 and 2 to find edge confidence: $C_{edge}$\;
        Equation 3 to find weighted edge confidence: $weighted\_C_{edge}$\;
        $CONF.append(weighted\_C_{edge})$\;
    }
    $F \gets extract\_edge\_connections(P)$\;
    \KwRet{$F, CONF$}
\end{algorithm2e}

\section{Results of Simulation}

Here we present each individual structural causal models generated through application of structure learning algorithms on the datasets. We have applied $SLA = \{sla_1, sla_2\}$ on dataset $D=\{d_1, d_2, d_3\}$.

\begin{figure}[htbp]
    \centering
    \includegraphics[width=\linewidth]{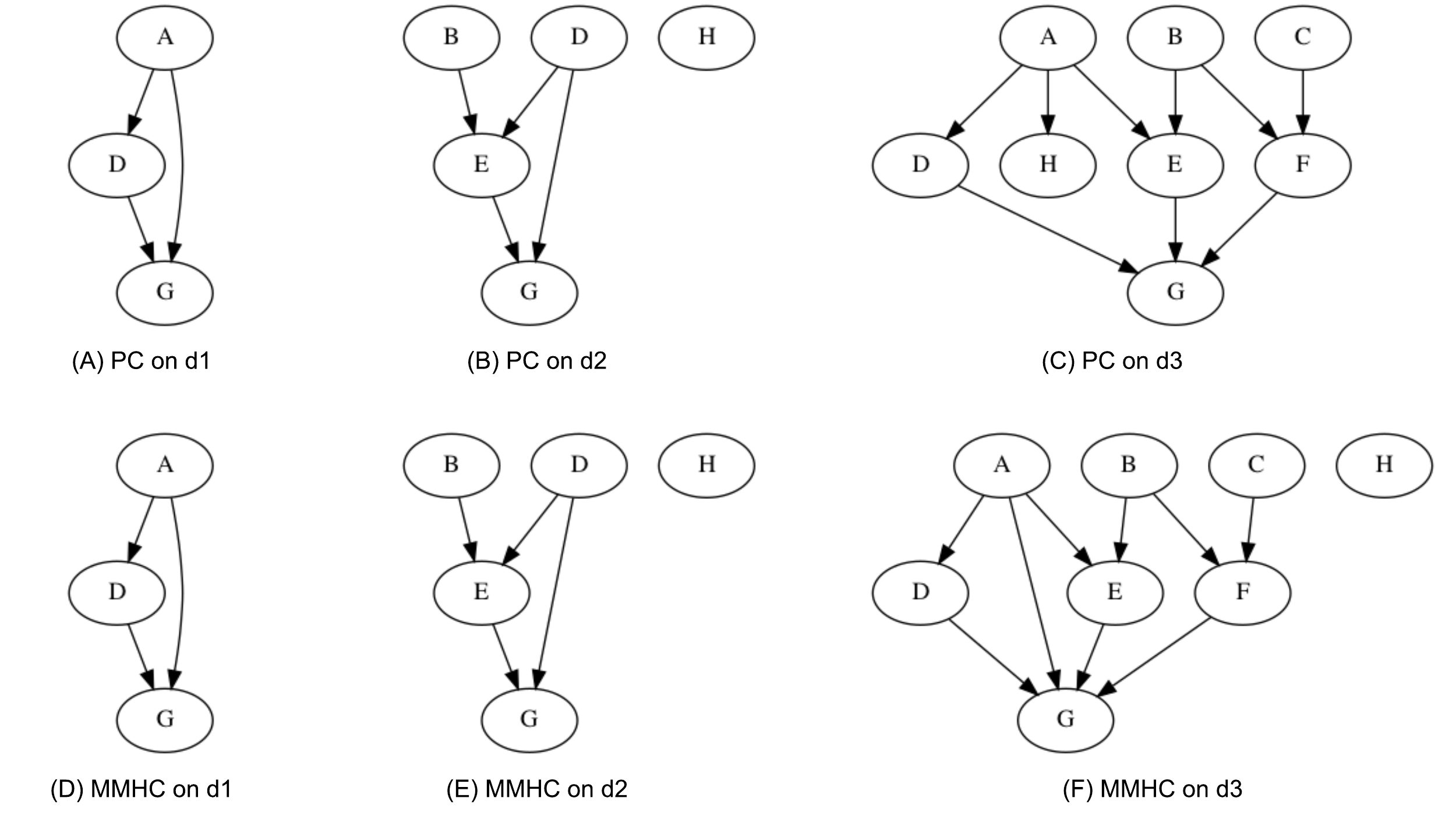}
    \caption{Structural Causal Models as outputs of Tier 2 in CKH}~\label{fig:sim2}
\end{figure}

The following table described the edge confidences, weighted edge confidences within each tier of CKH for this simulation.

\begin{landscape}
\begin{table}[ht]
  \caption{Summary of simulation results. Each tier has edge directions, edge confidence and weighted edge confidence. A positive value represents direction from first node to second node, and a negative value represents the opposite edge direction. Summarised output results are presented in the last two columns with edge directions and combined weighted edge confidences}
  \label{tab:result}
  \centering
  \resizebox{1.2\textwidth}{!}{
  \begin{tabular}{cccccccccccc}
    \toprule
    & \multicolumn{3}{c}{Tier 1 ($W_E = 0.2$)} & \multicolumn{3}{c}{Tier 2 ($W_D = 0.3$)} & \multicolumn{3}{c}{Tier 3 ($W_L = 0.5$)} & & \\
    \cmidrule(lr){2-4} \cmidrule(lr){5-7} \cmidrule(lr){8-10}
     & Edge & Edge & Weighted & Edge & Edge & Weighted & Edge & Edge & Weighted & Final & Combined \\
    Vars & Direction & Conf. & Edge Conf. & Direction & Conf. & Edge Conf. & Direction & Conf. & Edge Conf. & Direction & Conf. \\
    \midrule
    A, B & $A \dots B$ & 0.0 & 0.0 & $A \dots B$ & 0.0 & 0.0 & $A \dots B$ & 0.0 & 0.0 & $A \dots B$ & 0.0 \\
A, C & $A \dots C$ & 0.0 & 0.0 & $A \dots C$ & 0.0 & 0.0 & $A \dots C$ & 0.0 & 0.0 & $A \dots C$ & 0.0 \\
A, D & $A \dots D$ & 0.0 & 0.0 & $A \rightarrow D$ & 1.0 & 0.1116 & $A \dots D$ & 0.0 & 0.0 & $A \rightarrow D$ & 0.1116 \\
A, E & $A \dots E$ & 0.0 & 0.0 & $A \rightarrow E$ & 1.0 & 0.1116 & $A \dots E$ & 0.0 & 0.0 & $A \rightarrow E$ & 0.1116 \\
A, F & $A \dots F$ & 0.0 & 0.0 & $A \dots F$ & 0.0 & 0.0 & $A \dots F$ & 0.0 & 0.0 & $A \dots F$ & 0.0 \\
A, G & $A \dots G$ & 0.0 & 0.0 & $A \rightarrow G$ & 0.75 & 0.0837 & $A \dots G$ & 0.0 & 0.0 & $A \rightarrow G$ & 0.0837 \\
A, H & $A \dots H$ & 0.0 & 0.0 & $A \rightarrow H$ & 0.5 & 0.0558 & $A \dots H$ & 0.0 & 0.0 & $A \rightarrow H$ & 0.0558 \\
B, C & $B \dots C$ & 0.0 & 0.0 & $B \dots C$ & 0.0 & 0.0 & $B \dots C$ & 0.0 & 0.0 & $B \dots C$ & 0.0 \\
B, D & $B \dots D$ & 0.0 & 0.0 & $B \dots D$ & 0.0 & 0.0 & $B \dots D$ & 0.0 & 0.0 & $B \dots D$ & 0.0 \\
B, E & $B \rightarrow E$ & 1.0 & 0.2 & $B \rightarrow E$ & 1.0 & 0.1116 & $B \dots E$ & 0.0 & 0.0 & $B \rightarrow E$ & 0.3116 \\
B, F & $B \dots F$ & 0.0 & 0.0 & $B \rightarrow F$ & 1.0 & 0.1116 & $B \dots F$ & 0.0 & 0.0 & $B \rightarrow F$ & 0.1116 \\
B, G & $B \dots G$ & 0.0 & 0.0 & $B \dots G$ & 0.0 & 0.0 & $B \dots G$ & 0.0 & 0.0 & $B \dots G$ & 0.0 \\
B, H & $B \dots H$ & 0.0 & 0.0 & $B \dots H$ & 0.0 & 0.0 & $B \dots H$ & 0.0 & 0.0 & $B \dots H$ & 0.0 \\
C, D & $C \dots D$ & 0.0 & 0.0 & $C \dots D$ & 0.0 & 0.0 & $C \dots D$ & 0.0 & 0.0 & $C \dots D$ & 0.0 \\
C, E & $C \dots E$ & 0.0 & 0.0 & $C \dots E$ & 0.0 & 0.0 & $C \dots E$ & 0.0 & 0.0 & $C \dots E$ & 0.0 \\
C, F & $C \dots F$ & 0.0 & 0.0 & $C \rightarrow F$ & 1.0 & 0.1116 & $C \rightarrow F$ & 0.6667 & 0.0744 & $C \rightarrow F$ & 0.186 \\
C, G & $C \dots G$ & 0.0 & 0.0 & $C \dots G$ & 0.0 & 0.0 & $C \dots G$ & 0.0 & 0.0 & $C \dots G$ & 0.0 \\
C, H & $C \dots H$ & 0.0 & 0.0 & $C \dots H$ & 0.0 & 0.0 & $C \dots H$ & 0.0 & 0.0 & $C \dots H$ & 0.0 \\
D, E & $D \dots E$ & 0.0 & 0.0 & $D \rightarrow E$ & 0.5 & 0.0558 & $D \dots E$ & 0.0 & 0.0 & $D \rightarrow E$ & 0.0558 \\
D, F & $D \dots F$ & 0.0 & 0.0 & $D \dots F$ & 0.0 & 0.0 & $D \dots F$ & 0.0 & 0.0 & $D \dots F$ & 0.0 \\
D, G & $D \rightarrow G$ & 1.0 & 0.2 & $D \rightarrow G$ & 1.0 & 0.1116 & $D \rightarrow G$ & 1.0 & 0.1116 & $D \rightarrow G$ & 0.4232 \\
D, H & $D \dots H$ & 0.0 & 0.0 & $D \dots H$ & 0.0 & 0.0 & $D \dots H$ & 0.0 & 0.0 & $D \dots H$ & 0.0 \\
E, F & $E \dots F$ & 0.0 & 0.0 & $E \dots F$ & 0.0 & 0.0 & $E \dots F$ & 0.0 & 0.0 & $E \dots F$ & 0.0 \\
E, G & $E \dots G$ & 0.0 & 0.0 & $E \rightarrow G$ & 1.0 & 0.1116 & $E \rightarrow G$ & 0.6667 & 0.0744 & $E \rightarrow G$ & 0.186 \\
E, H & $E \dots H$ & 0.0 & 0.0 & $E \dots H$ & 0.0 & 0.0 & $E \dots H$ & 0.0 & 0.0 & $E \dots H$ & 0.0 \\
F, G & $F \dots G$ & 0.0 & 0.0 & $F \rightarrow G$ & 1.0 & 0.1116 & $F \dots G$ & 0.0 & 0.0 & $F \rightarrow G$ & 0.1116 \\
F, H & $F \dots H$ & 0.0 & 0.0 & $F \dots H$ & 0.0 & 0.0 & $F \dots H$ & 0.0 & 0.0 & $F \dots H$ & 0.0 \\
G, H & $G \dots H$ & 0.0 & 0.0 & $G \dots H$ & 0.0 & 0.0 & $G \dots H$ & 0.0 & 0.0 & $G \dots H$ & 0.0 \\
    \bottomrule
  \end{tabular}}
\end{table}
\end{landscape}

\section{Extended Discussion}

\paragraph{Rationale with the Theory of SCM}
Finding the right and the most-compatible causal model with the underlying data generating mechanism is critical for estimating causal effects through rules of do-calculus. This is a challenging research problem since data itself cannot differentiate between SCM within a Markov equivalence class. It has been proven in a recent seminal work \citep{bareinboim2016causal} that additional complementary causal knowledge is needed along with data to derive a SCM and the proposed methodology can be a strong tool to aid in that process. Our proposed methodology uses a systematic approach to incrementally derive the SCM with appropriated scoring for different levels of evidence and can generate high accuracy even in the presence of contradictory causal connections. The algorithm would be beneficial for applied causal inference researchers, specially in epidemiology, medicine and social sciences. Proposed CKH-driven algorithm effectively estimates all necessary components of a computed SCM, $U, V, F$ from all tiers of causal knowledge and $P(u)$ from Tier 2 (data) only. However, it also produces weighted edge confidences $CONF$, which is a key contribution of this algorithm. The edge confidence signifies the strength or confidence of information we have on the specific edge, however it does not state anything about the strength of causal relationships between variables or parameters of functions $F$. 

\paragraph{Completeness of the Derived Causal Model}
The causal model estimated through CKH algorithm is built through encoding of causal knowledge sources iteratively and thus holds a summary of causal information from all possible and relevant sources. The rationale for its completeness is that, $\langle U, V, F, P(u) \rangle$, values of four (4) components of the structural causal model derived with the proposed CKH-driven algorithm, is curated from all the tiers of CKH \textit{(and thus contains all necessary and relevant information needed to generate a complete outcome)}. This collaborated information is also weighted appropriately based on the significance and the impact of causal knowledge sources. Another key argument for completeness of the derived causal model comes from the tier weights and their usage. Till now, we have used an increasing weight for the three tiers ($W_E < W_D < W_L$) for a well-researched problem domain, with sufficient experts, data and publications available on the domain specific problem of interest. For an evolving problem domain (e.g., COVID-19 crisis), where we do not have an abundance of well-established peer-reviewed literature, we can alter and adjust the tier weights as fit for the problem at hand. For example, in estimating the effect of a specific old drug in treatment of COVID-19 patients, we would have more weight on Tier 2 (data, $CK_D$) compared to Tier 3 (literature, $CK_L$), simply because of lack of strong evidence from literature and might use an alternative variation of tier weights hierarchy ($W_E \leq W_D > W_L$). For such reasons, we conjecture that CKH provides a fundamental, necessary, and sufficient building mechanism for constructing structural causal models for a problem domain, given causal knowledge from a varying sources. A rigorous proof for the completeness is still under investigation.



\paragraph{Application in specific fields of Science}
Proposed CKH-driven causal model generation has high impact for specific fields of science, and especially in health science. Identifying the cause for an outcome and quantifying the causal effect is of high importance in health science and epidemiology. An ongoing work is aiming to derive a SCM for the treatment of delirium patients in the ICU \citep{bikak2019outcomes} based on the CKH. Other than that, CKH has implications in other branches of science, where the notion of causality is critical, such as, sociology and finance.